# Policy Gradient Methods for Reinforcement Learning with Function Approximation and Action-Dependent Baselines

Philip S. Thomas  
Carnegie Mellon University

Emma Brunskill  
Stanford University

We show how an action-dependent baseline can be used by the policy gradient theorem using function approximation, originally presented with action-independent baselines by Sutton et al. (2000).

## 1 Notation and Background

We assume that the reader is familiar with the seminal paper by Sutton et al. (2000), which shows how the policy gradient theorem (Baxter and Bartlett, 1999) can be extended to include function approximation and action-independent baselines. Our paper is intended to be read immediately after reviewing Section 3 of the paper by Sutton et al. (2000). Although here we adopt the episodic setting, the extension of this work to the average reward setting is straightforward. We use the notational standard MDPNv1 (Thomas and Okal, 2015) and the following additional definitions, where expectations conditioned on $\theta$ denote that actions, $A_t$, are sampled from $\pi(S_t, \cdot, \theta)$ unless otherwise specified:

$$\text{Objective function:} \quad \rho(\theta) := \mathbf{E}\left[\sum_{t=0}^{\infty} \gamma^t R_t \,\Big|\, \theta\right]$$

$$\text{State value function:} \quad v_\theta(s) := \mathbf{E}\left[\sum_{t=0}^{\infty} \gamma^t R_t \,\Big|\, S_0 = s, \theta\right]$$

$$\text{State-action value function:} \quad q_\theta(s,a) := \mathbf{E}\left[\sum_{t=0}^{\infty} \gamma^t R_t \,\Big|\, S_0 = s, A_0 = a, \theta\right]$$

$$\text{Discounted state distribution:} \quad d_\theta(s) := \sum_{t=0}^{\infty} \gamma^t \Pr(S_t = s | \theta)$$

$$\text{Compatible function approximator:} \quad f_w(s,a) := w^\intercal \frac{\partial}{\partial \theta} \ln(\pi(s,a,\theta))$$

$$\text{Loss function:} \quad L(w) := \frac{1}{2} \sum_s d_\theta(s) \sum_a \pi(s,a,\theta)\Big(f_w(s,a) - q_\theta(s,a)\Big)^2$$

$$\text{Critical } q\text{-parameters:} \quad w^\star \in \left\{w \in \mathbb{R}^{n_\theta} : \frac{\partial}{\partial w} L(w) = 0\right\}.$$

## 2 The Policy Gradient Theorem With Function Approximation and Action-Independent Baselines

The policy gradient theorem with function approximation and action-**independent** baselines, as presented by Sutton et al. (2000), states that for all $b : \mathcal{S} \to \mathbb{R}$,

$$\nabla \rho(\theta) = \sum_{s \in \mathcal{S}} d_\theta(s) \sum_{a \in \mathcal{A}} \pi(s,a,\theta)\left(f_{w^\star}(s,a) - b(s)\right) \frac{\partial}{\partial \theta} \ln(\pi(s,a,\theta)).$$

The action-independent baseline, $b$, does not add bias because:

$$\sum_{s \in \mathcal{S}} d_\theta(s) \sum_{a \in \mathcal{A}} \pi(s,a,\theta) b(s) \frac{\partial}{\partial \theta} \ln(\pi(s,a,\theta)) = \sum_{s \in \mathcal{S}} d_\theta(s) b(s) \overbrace{\frac{\partial}{\partial \theta} \underbrace{\sum_{a \in \mathcal{A}} \pi(s,a,\theta)}_{=1}}^{=0} = 0.$$

However, introducing an action-dependent baseline, $b : \mathcal{S} \times \mathcal{A} \to \mathbb{R}$, can introduce bias because $b(s,a)$ cannot be pulled out of the sum over actions, and so the derivative of the sum over actions does not necessarily evaluate to zero.

# 3 The Policy Gradient Theorem With Function Approximation and Action-Dependent Baselines

We will show how an action-dependent baseline, $b : \mathcal{S} \times \mathcal{A} \to \mathbb{R}$, can be incorporated into the policy gradient theorem with function approximation without introducing bias. To do so, we define a different loss function, $\tilde{L}$:

$$\tilde{L}(w) \coloneqq \frac{1}{2} \sum_s d_\theta(s) \sum_a \pi(s,a,\theta) \Big(f_w(s,a) - (q_\theta(s,a) - b(s,a))\Big)^2.$$

Intuitively, $L$ is minimized by weights, $w$, that cause $f_w(s,a)$ to approximate the state-action value function, $q_\theta(s,a)$, while $\tilde{L}$ is minimized by weights that cause $f_w(s,a)$ to approximate the residual error after the baseline is subtracted from the state-action value function. Let

$$\widetilde{w}^\star \in \left\{ w \in \mathbb{R}^{n_\theta} : \frac{\partial}{\partial w} \tilde{L}(w) = 0 \right\}.$$

In Theorem 1 we show that using an action-dependent baseline does not introduce bias if the compatible function approximator is trained to estimate the residual rather than the state-action value function, i.e., if we use $\widetilde{w}^\star$ rather than $w^\star$:

**Theorem 1** (Policy gradient theorem with function approximation and action-dependent baseline). *For any action-dependent baseline, $b : \mathcal{S} \times \mathcal{A} \to \mathbb{R}$,*

$$\nabla \rho(\theta) = \sum_s d_\theta(s) \sum_a \pi(s,a,\theta) \Big(f_{\widetilde{w}^\star}(s,a) + b(s,a)\Big) \frac{\partial}{\partial \theta} \ln(\pi(s,a,\theta)).$$

*Proof.* By the definition of $\widetilde{w}^\star$ we have that:

$$\frac{\partial}{\partial \widetilde{w}^\star} \tilde{L}(\widetilde{w}^\star) = 0$$

$$\sum_s d_\theta(s) \sum_a \pi(s,a,\theta) \Big(f_{\widetilde{w}^\star}(s,a) - (q_\theta(s,a) - b(s,a))\Big) \underbrace{\frac{\partial}{\partial \widetilde{w}^\star} f_{\widetilde{w}^\star}(s,a)}_{= \frac{\partial}{\partial \theta} \ln(\pi(s,a,\theta))} = 0$$

$$\sum_s d_\theta(s) \sum_a \pi(s,a,\theta) \Big(f_{\widetilde{w}^\star}(s,a) + b(s,a)\Big) \frac{\partial}{\partial \theta} \ln(\pi(s,a,\theta)) = \underbrace{\sum_s d_\theta(s) \sum_a \pi(s,a,\theta) q_\theta(s,a) \frac{\partial}{\partial \theta} \ln(\pi(s,a,\theta))}_{= \nabla \rho(\theta)}.$$

$\square$

# 4 Discussion

Theorem 1 has interesting ramifications—it suggests several new policy gradient algorithms. First, one might define $b(s,a)$ to be an estimate of $q_\theta(s,a)$ constructed from an approximate MDP model, for example, one built from expert knowledge. Alternatively, one might estimate $b(s,a)$ from data. In this latter setting, one might parameterize $b$ by a vector, $x$, and tune $x$ to make $b_x$ approximate $q_\theta$. This could be done prior to approximating $\widetilde{w}^\star$, or could be done simultaneously by viewing $\hat{q}_{w,x}(s,a) \coloneqq f_w(s,a) + b_x(s,a)$ as a new function approximator with parameter vector $(w^\intercal, x^\intercal)^\intercal$, and searching for weights that make $\hat{q}_{w,x}$ approximate $q_\theta$.

## Acknowledgements

The research reported here was supported in part by a NSF CAREER grant and the Institute of Education Sciences, U.S. Department of Education, through Grant R305A130215 to Carnegie Mellon University. The opinions expressed are those of the authors and do not represent views of the Institute or the U.S. Dept. of Education or NSF.